\pdfoutput=1
\documentclass[conference]{IEEEtran}

\usepackage[utf8]{inputenc}
\usepackage[english]{babel}
\usepackage{caption}
\usepackage{subfig}
\usepackage[backend=bibtex]{biblatex}
\usepackage{color}
\usepackage{graphicx}
\usepackage{pdfpages}
\bibliography{refs}
\begin{document}

\title{Autonomous Shuttles for Last-Mile Connectivity}

\author{\IEEEauthorblockN{Garrison Neel, Amir Darwesh, Quang Le, Srikanth Saripalli}
\IEEEauthorblockA{
Dept. of Mechanical Engineering\\
Texas A\&M University\\
College Station, Texas 77840\\
Email: \{gneel, adarwesh, qle, ssaripalli\}@tamu.edu}
}
\maketitle

\begin{abstract}
This paper describes an autonomous shuttle which targets providing last-mile transportation. 
Often, this involves operation in crowded areas with high levels of pedestrian traffic, and little to no lane markings or traffic control.
We aim to create a functional shuttle to be improved upon in the future as new robust solutions are developed to replace the current components.
An initial implementation of such a shuttle presented, detailing the overall architecture, controller structure, waypoint following, obstacle detection and avoidance, LiDAR based sign detection, and pedestrian communication.
The performance of each component is evaluated, and future improvements are discussed.

\end{abstract}

\section{Introduction}

The last-mile refers to the final portion of a journey, between a transportation hub and the final destination. Often, this portion of a trip contributes disproportionately to the overall length of a trip \cite{last_mile}. Autonomous shuttles are a convenient solution to the last mile problem. Because of their ability to drive on pedestrian pathways, an shuttle can deliver passengers directly to their destination. Companies such as May Mobility \cite{etherington_2018} and Optimus Ride \cite{harris_2016} are implementing shuttles in urban areas toward this end. 

Autonomous shuttles have many benefits and challenges when compared to traditional autonomous vehicles. Because they can drive on pedestrian paths, they can reach many areas cars cannot, such as the inner spaces of a college campus. These paths, however, are not easily navigable. They are often narrow, with no lane markings, and crowded with a mix of pedestrian and bike traffic. Previous studies, such as Chong et al. \cite{Chong_2011} have emphasized the importance of good obstacle detection, and localization in these environments. Pendleton et al. \cite{Pendleton_2017} reviews alternative options in each area fundamental to autonomous vehicle operation.

The goal of this project is to implement an autonomous shuttle which can operate in these conditions, and which is upgradeable as new sensors are available and new software is written. The shuttle presented is capable of following prescribed paths, stopping for and yielding to pedestrian and bike traffic, and communicating intent to pedestrians. It is a Polaris GEM e4, retrofitted to be drive-by-wire, purchased from AutonomouStuff \cite{astuff}.

\section{System Architecture}

\begin{figure}[h!]
\includegraphics[width=0.5\textwidth]{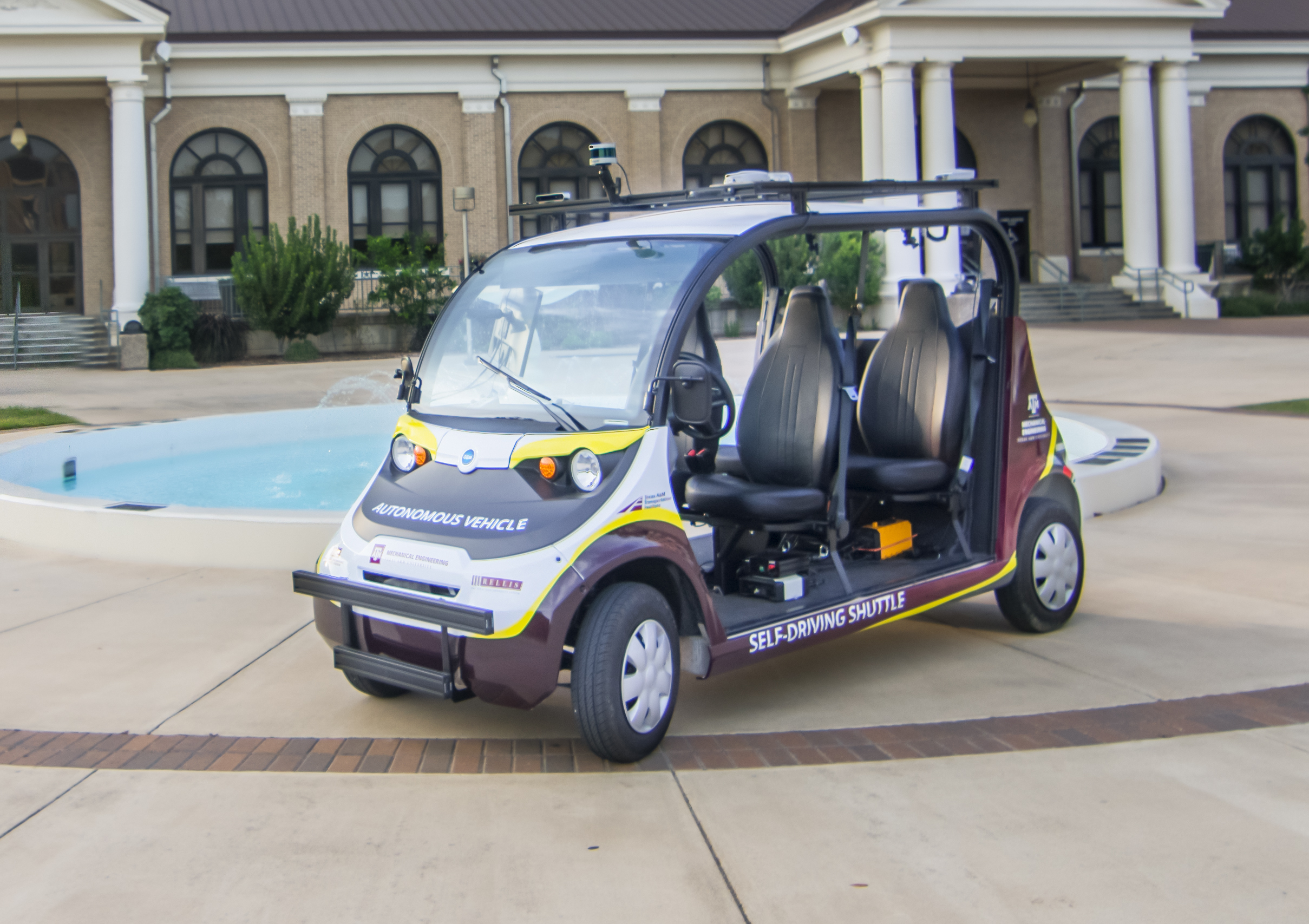}
\caption{The Polaris shuttle equipped to drive autonomously}
\label{photo}
\end{figure}

The vehicle used is a Polaris GEM e4 purchased from AutonomouStuff \cite{astuff}, pictured in Figure \ref{photo}. AutonomouStuff modifies the vehicle with drive-by-wire functionality using a module referred to as PACMod \cite{pacmod}. An on-board computer uses CAN to send commands and read sensor data from PACMod. A Velodyne VLP-16 \cite{vlp-16} LiDAR is mounted to the roof at the front of the vehicle for use in obstacle and sign detection, and a Vectornav VN-300 \cite{vn-300} GPS/IMU is used for localization. Additional signs and speakers mounted at the front allow it to communicate with pedestrians. 80/20 rails on the roof and bumpers provide ample space to mount additional sensors. The on-board computer and all sensors which require 120V power, are powered by auxiliary batteries mounted under the front seats.

\begin{figure}[h!]
\includegraphics[width=0.49\textwidth]{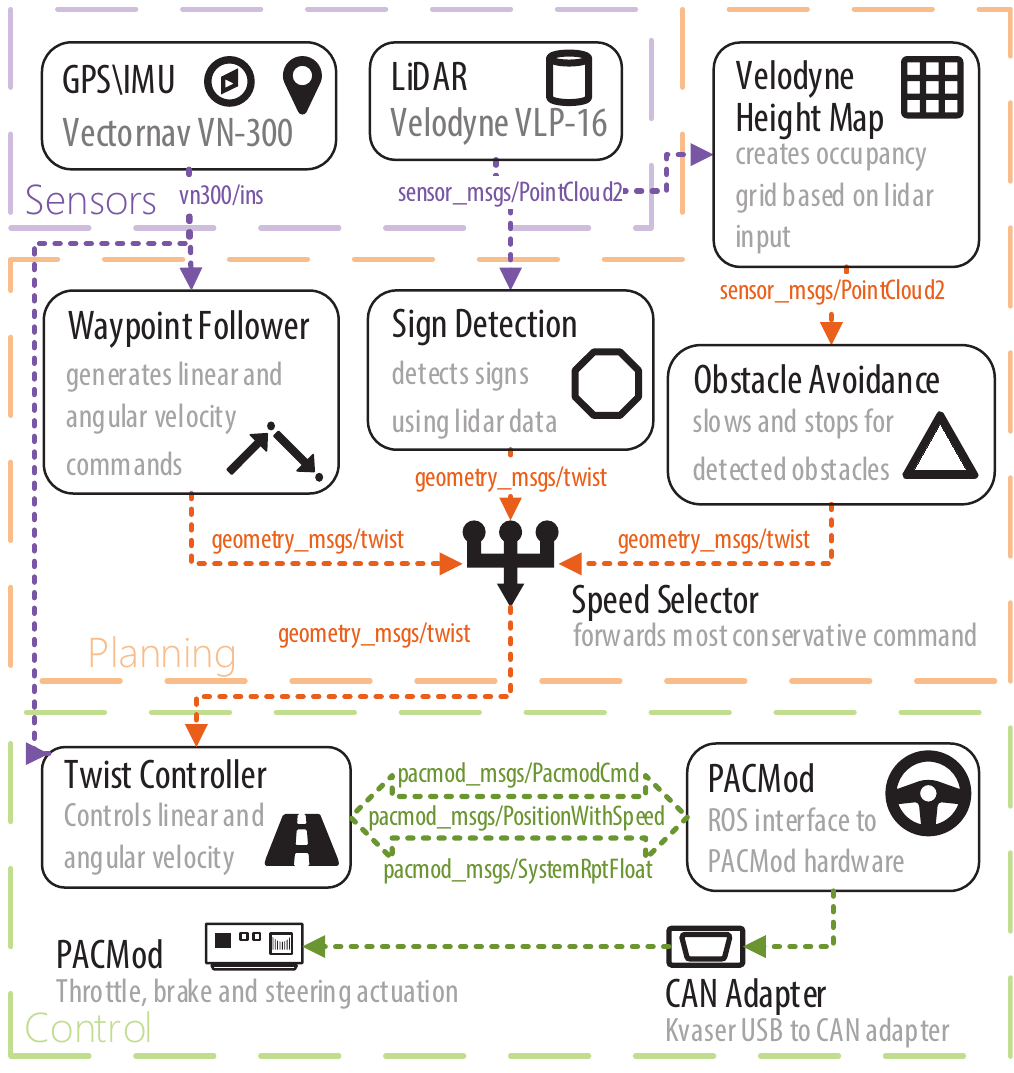}
\caption{ROS system architecture diagram of the shuttle showing relationship between ROS packages}
\label{system_arch}
\end{figure}

All control and sensing software is integrated using the Robot Operating System (ROS) \cite{ros.org}. ROS is used to facilitate communication between the individual components of the system. Messages are passed betweenn modules using a publisher/subscriber architecture. Figure \ref{system_arch} shows the overall ROS system architecture. The waypoint follower takes input from the GPS to calculate a desired linear and angular velocity command. This commands are subscribed to by the obstacle avoidance and sign detection packages. These packages update the command, and publish them again to the core control package. This package selects the lowest command velocity from all of the inputs and publishes it to the twist controller. The twist controller then generates throttle, brake and steering commands, and publishes them to PACMod in order to achieve the desired linear and angular velocity.

\section{Controller Structure}

The twist controller shown in Figure \ref{system_arch} is the linear and angular velocity controller for the shuttle. It takes target velocities with acceleration limits as inputs and outputs steering and throttle or brake values to the PACMod driver.
\subsection{Linear Velocity}
\begin{figure}[h!]
\frame{\includegraphics[width=0.49\textwidth]{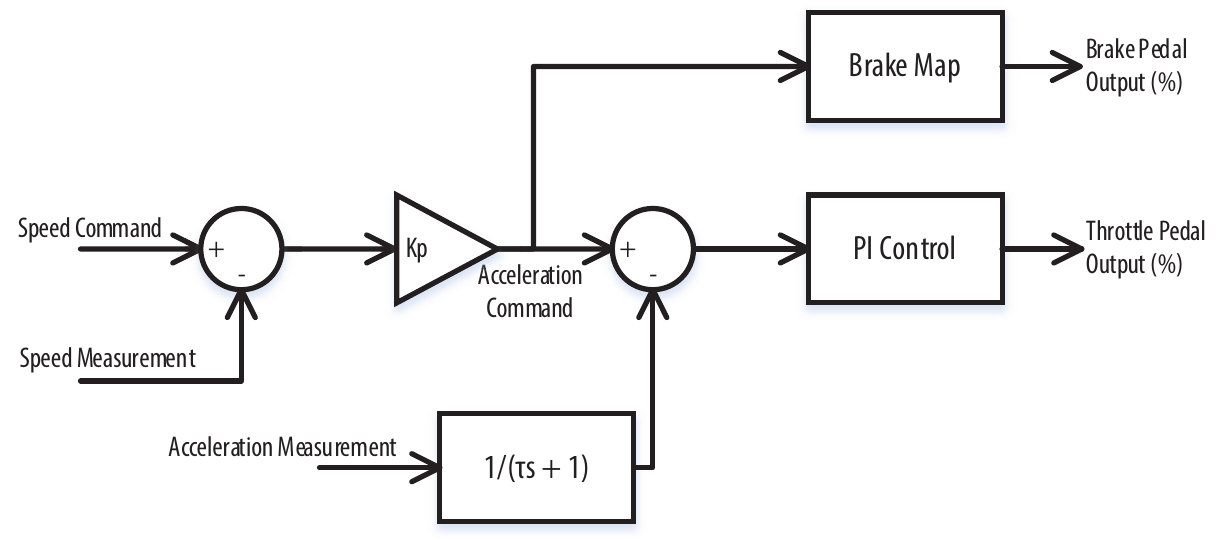}}
\caption{Speed controller block diagram}
\label{speed_controller}
\end{figure}

\begin{figure}[h!]
\includegraphics[width=0.5\textwidth]{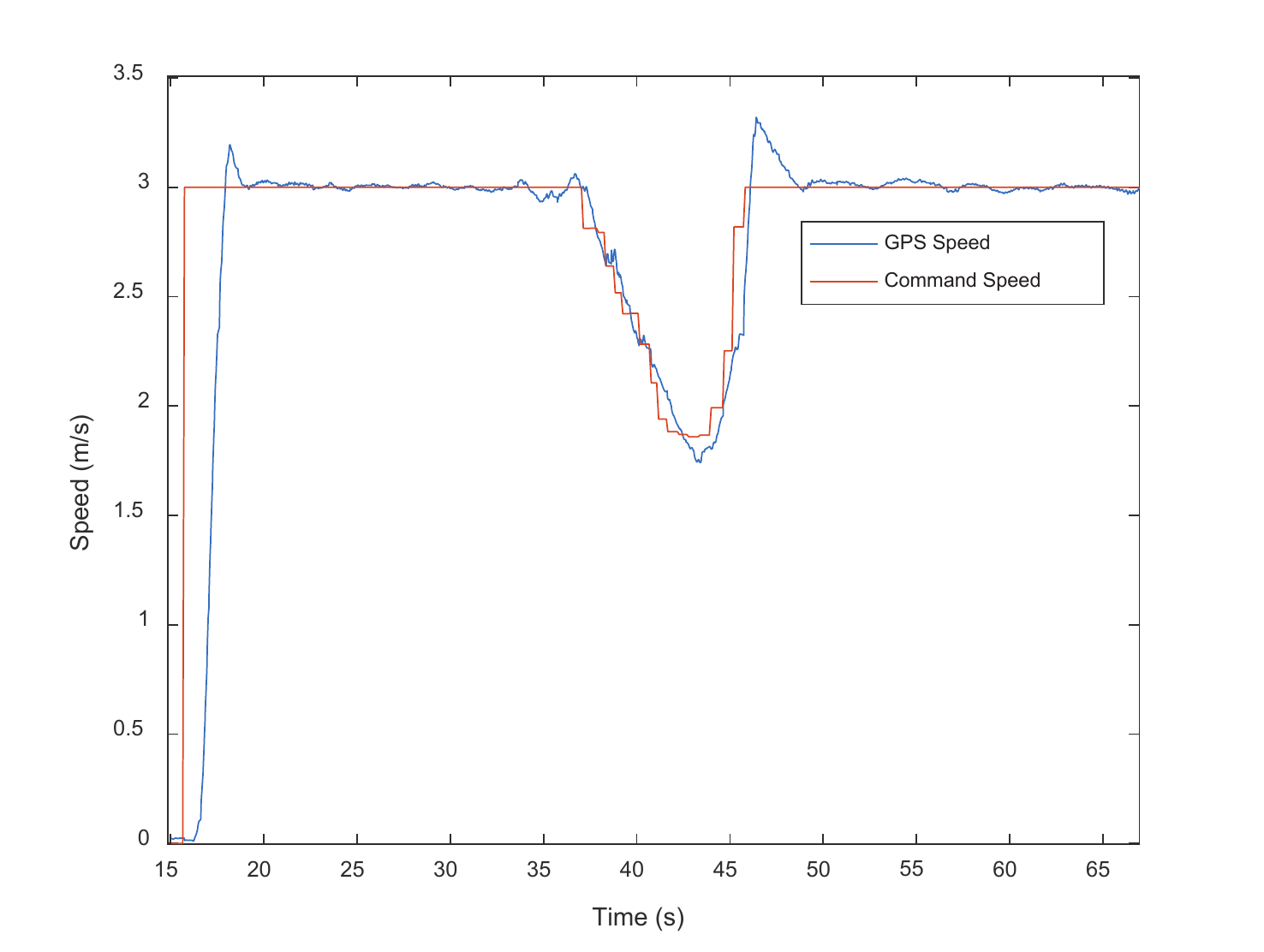}
\caption{A plot showing the forward speed controller performance}
\label{speed_control}
\end{figure}

The linear velocity controller is shown in Figure \ref{speed_controller}. Command velocity is input, and velocity feedback is read from the GPS. The acceleration command coming from the speed error proportional gain is limited to maintain a comfortable ride. Acceleration feedback comes from the on-board IMU, and is filtered to remove noise using $\tau = 2$, where $\tau$ is the time constant of the low-pass filter. When the acceleration error is positive, a PI controller is used to generate throttle values. The throttle output is filtered to eliminate jerk and smooth the ride in favor of strict controller performance. For negative acceleration error, an open-loop controller is used due to the nonlinear nature of the braking. The experimentally determined lookup equation for braking it $b = 0.28\times\log{(a_{cmd})}+0.90$ where $a_{cmd}$ is the acceleration command, and $b$, the brake value, is limited to be between 0 and 1. %
Figure \ref{speed_control} shows the speed controller performance.


\subsection{Angular Velocity}
\begin{figure}[h!]
\includegraphics[width=0.5\textwidth]{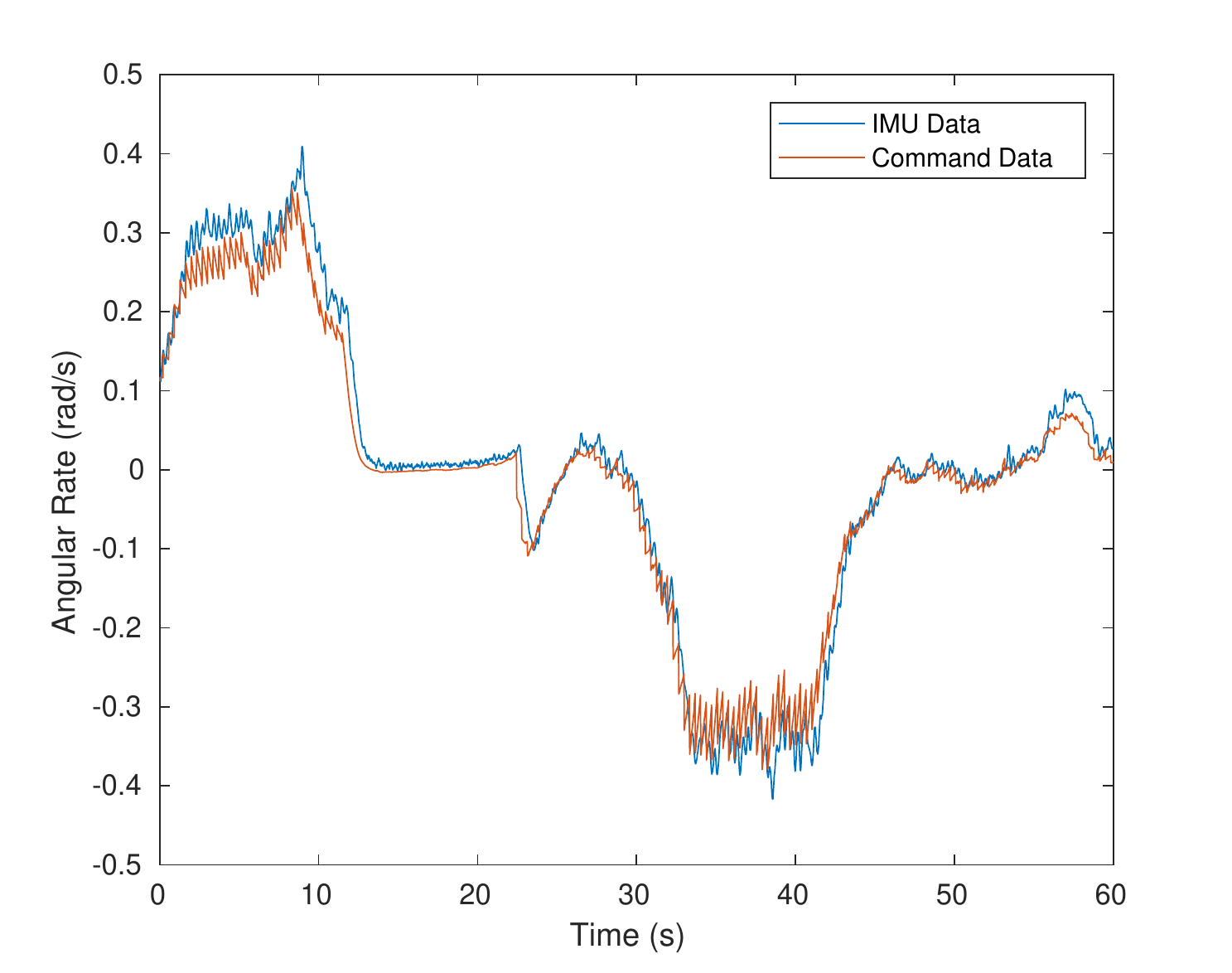}
\caption{A plot showing the angular speed controller performance}
\label{angular_velocity_control}
\end{figure}

Angular velocity is controlled using a bicycle model~\cite{kong_2015} to calculate the required steering based on the steering ratio, wheelbase, and forward velocity of the shuttle. Figure \ref{angular_velocity_control} shows how the measured angular velocity tracks the commanded angular velocity. 

\section{Waypoint Follower}
A waypoint following controller was implemented, giving the shuttle the ability to follow waypoint lists in series. Waypoint files consist of a series of triplets listing the latitude, longitude and speed of each waypoint. The controller uses the error in the shuttle's heading relative to the targeted waypoint to command an angular velocity. The commanded angular velocity is $\omega_c = K_p\times\theta_{error}$, where $K_p$ is a controller gain, $\theta_{error}$ is the difference between the current shuttle heading, and the heading of the target. 
Speed is read from the waypoint file, and the pair is then passed to the twist controller. The waypoints are targeted sequentially beginning with the first waypoint in the list, and the next waypoint is targeted once the golf-cart is within 2m of the current target waypoint.

Waypoint paths are created by driving the vehicle along the desired path and recording GPS position and speed at intervals of 1m. Curvature of the path is used to limit the speed included in the waypoint file before the files are generated. Approximate radius is calculated from the relationship $r = v\times\omega$, where $v$ and $\omega$ are the recorded linear and angular velocity of the shuttle while driving the desired path. This radius is then used to limit lateral acceleration to 0.5 $m/s^2$. $v_{max} = \sqrt[]{0.5\times r}$ where $r$ is the calculated radius of the path at the waypoint. Since the speed is included in the waypoint triplet, a separate appropriately named file is generated for each desired speed.

\begin{figure}[h!]
\includegraphics[width=0.5\textwidth]{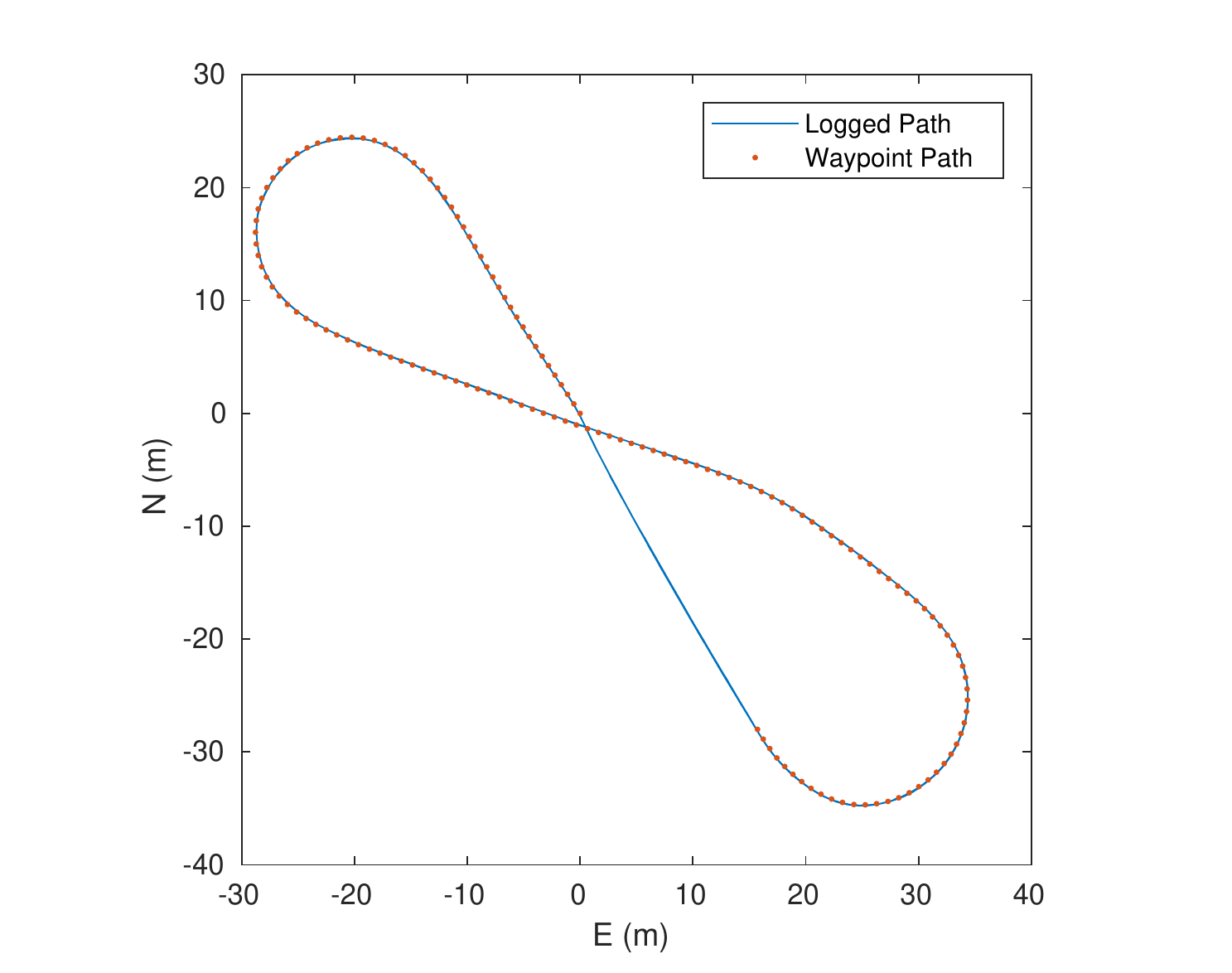}
\caption{A plot showing a figure-8 shaped waypoint path in red, with the logged path of the shuttle shown in blue}
\label{waypoint_path}
\end{figure}

\begin{figure}[h!]

\includegraphics[width=0.5\textwidth]{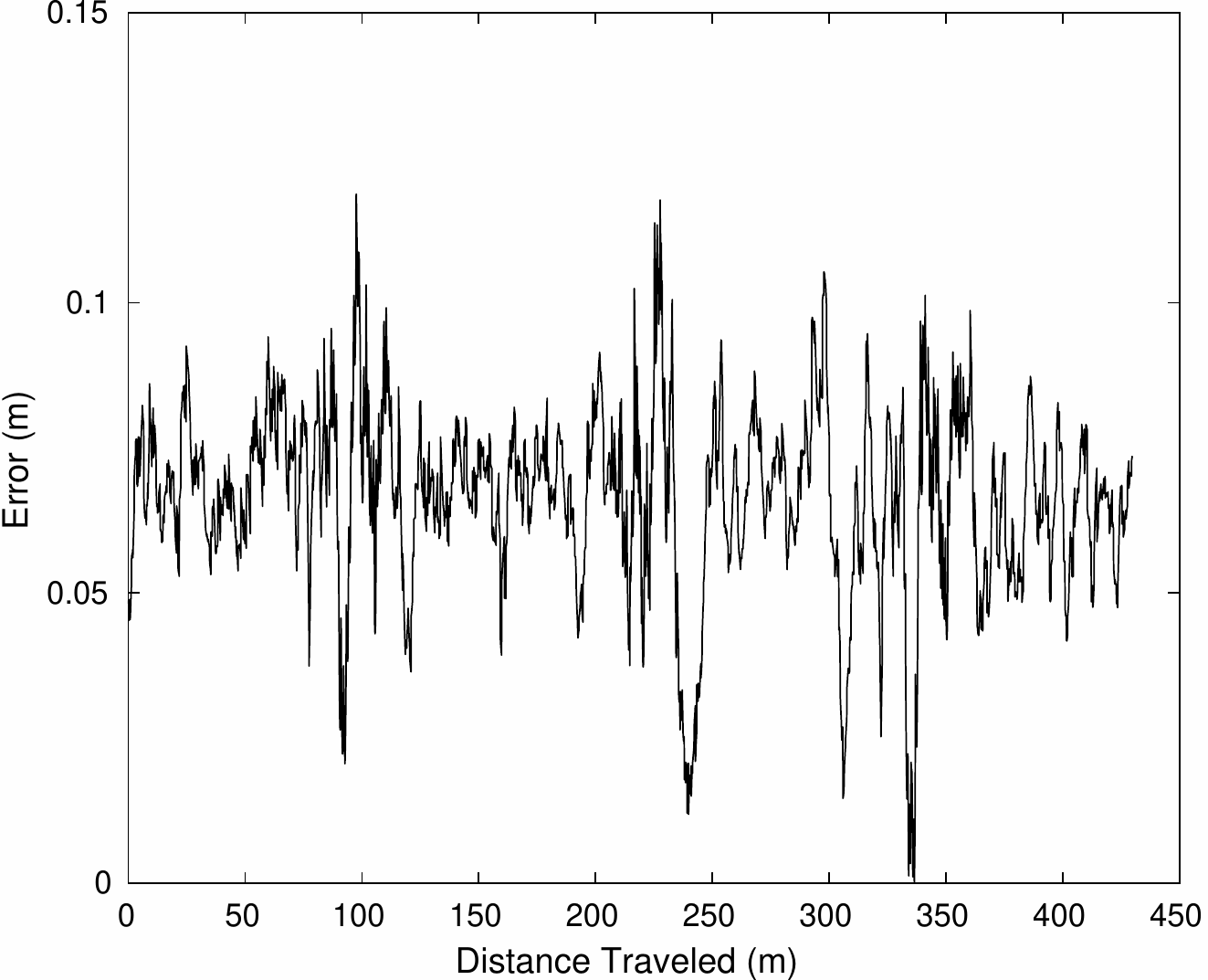}
\caption{A plot of absolute value of the cross-track error while following the figure-8 path shown in  figure \ref{waypoint_path} }
\label{crosstrack_error}
\end{figure}

Figure \ref{crosstrack_error} is a plot of the absolute value of cross-track error while following the path shown in Figure \ref{waypoint_path} at 3 m/s. As shown in the plot, peak cross-track error is 12 cm. The average error is relatively high, suggesting a small constant heading error. Results could be improved by adding a constant offset correcting the heading, or integral term to the waypoint controller. 


\section{Obstacle Detection}

Obstacle detection is done by building an occupancy grid of the surroundings. The occupancy grid is populated using a modified version of velodyne\_height\_map~\cite{velodyne_heightmap}. The modifications include removal of points above the shuttle, and setting the grid to 0.25 m with a height threshold of 0.07 m, in order to reliably detect pedestrian sized objects using the VLP-16 LiDAR. The current obstacle detection is not a replacement for a safety driver, and the shuttle is never operated autonomously without a human safety driver able to take manual control at any time.

\begin{figure}[h!]
\frame{\includegraphics[width=0.49\textwidth]{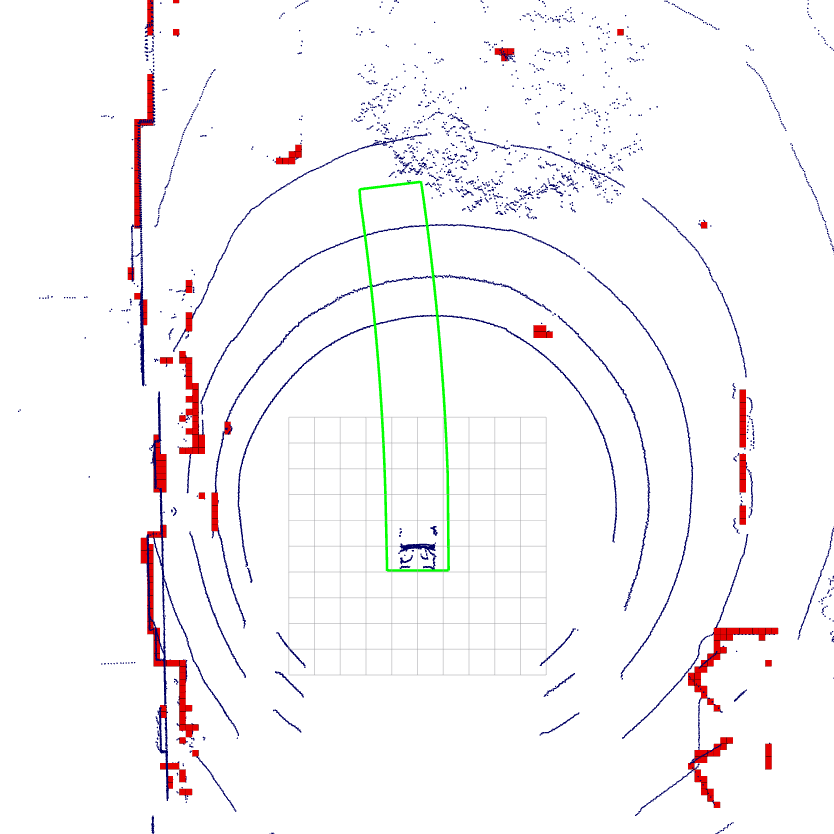}}
\caption{A visualization of the obstacle detection. This top-down view (with grid centered over the shuttle) shows the LiDAR point cloud in blue, detected obstacles in red, and the path of the shuttle is outlined in green.}
\label{obstacle_avoidance}
\end{figure}

The expected path of the shuttle is calculated from the current steering angle using a bicycle model~\cite{kong_2015}, and widened to cover the area the cart will traverse, plus some clearance. This calculated region and the obstacles detected are shown in Figure \ref{obstacle_avoidance}. From the detected obstacles, the closest along the path is selected and used to alter the twist message coming from the waypoint follower. Since there is rarely room for navigation around obstacles, the obstacle avoidance package only modifies the speed, reducing it based on the distance to the closest obstacle. The reduced velocity is $v = \frac{d}{5}-1$ where $d$ is the distance to the detected obstacle. Once the obstacle is within 5 m, the commanded speed is 0 m/s at the maximum deceleration rate, in order to stop the cart as quickly as possible.

From testing, the shuttle takes 1.6m, and 0.8s to stop completely under full braking from a speed of 3 m/s. Assuming a pedestrian walking speed of 1.4 m/s~\cite{mutcd_2012}, this means a side clearance of 1.15 m is required to prevent collision with a pedestrian walking perpendicular to the path. If the vehicle speed is increased to 5 m/s, the required clearance increases to 1.7 m. Typically, in the areas where the shuttle is operated, a smaller clearance is required to allow for passage between stationary obstacles. Additionally, the traffic in these areas is largely in the same or opposite direction of the path, rarely perpendicular, so a narrower clearance is sufficient, as long as a safety driver is present. Since full braking is applied for obstacles within 5 m, the calculated upper limit for oncoming traffic is 3.4 m/s for operation at 3 m/s. However, because the path ahead is checked for 15 m, there is no issue slowing and stopping for faster traffic in practice. %

All of the previous discussion assumes reliable obstacle detection. A height map is not well suited for detecting obstacles which are not well-covered by the LiDAR. Moving obstacles are often missed temporarily as they pass between grid squares. Close obstacles which have relatively flat tops, such as the roof of a car or the bed of a truck, are not detected by this algorithm. In addition to software limitations, the roof-mounted LiDAR has a large cone-shaped blind spot beneath it with a base radius of approximately 7.5 m. This means obstacles below 1.3 m in height cannot be detected if they are within 1 m of the front bumper of the shuttle. Due to the limitations of the obstacle detection, the shuttle is always operated with a safety driver. Future implementations of obstacle avoidance will use ground plane segmentation  and point clustering to detect obstacles and track them to allow operation in areas with faster-moving traffic. Additional hardware will be mounted to eliminate the blind spots of the roof-mounted LiDAR. %

\section{LIDAR Based Sign Detection}

\begin{figure}[!h]%
    \centering
    \subfloat[Type I]{{\includegraphics[width=0.1\textwidth]{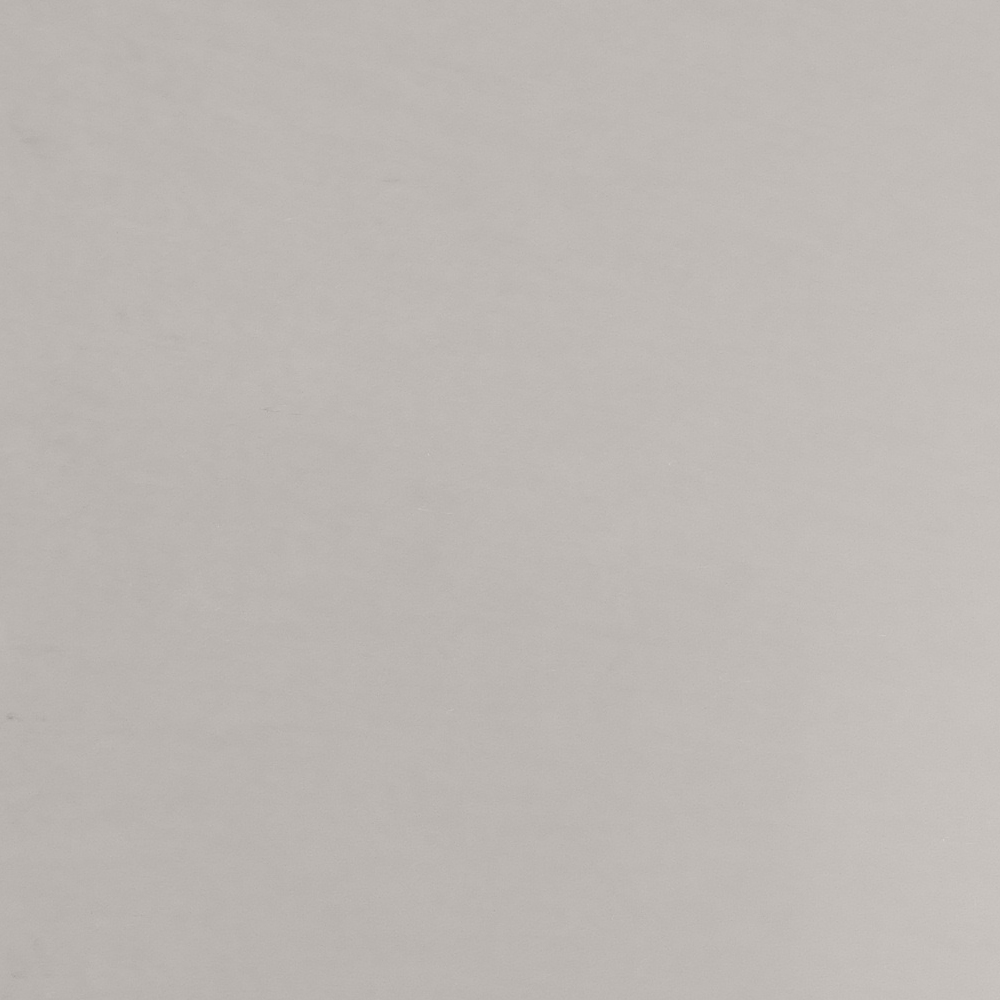} }}%
    \subfloat[Type II]{{\includegraphics[width=0.1\textwidth]{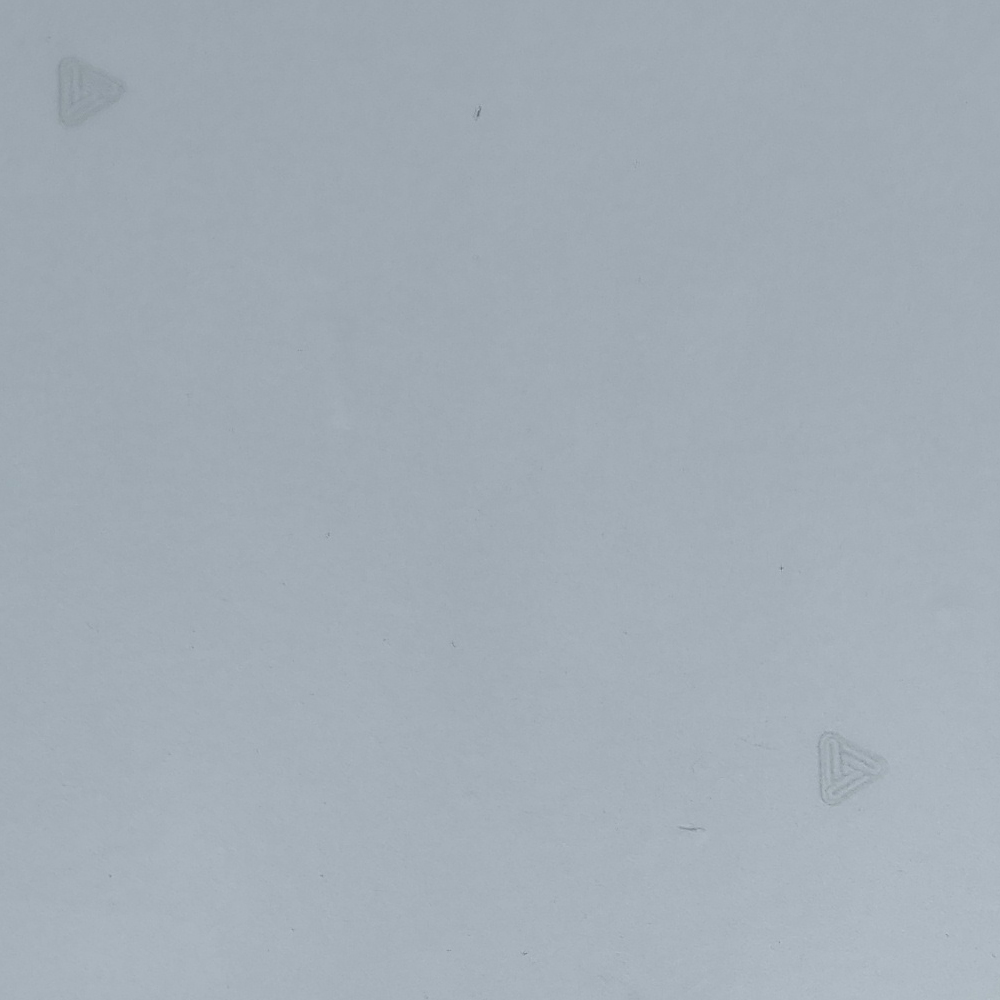} }}%
    \subfloat[Type III]{{\includegraphics[width=0.1\textwidth]{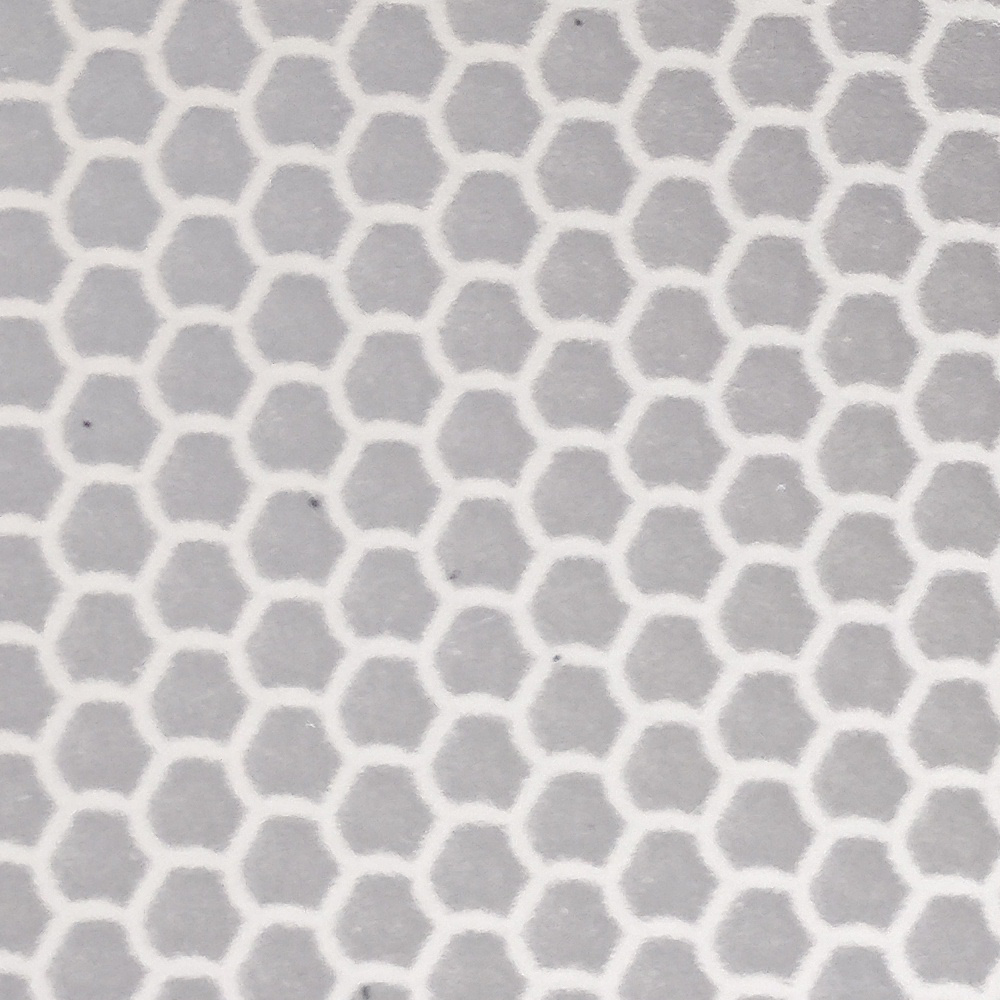} }}%
    \subfloat[Type IV]{{\includegraphics[width=0.1\textwidth]{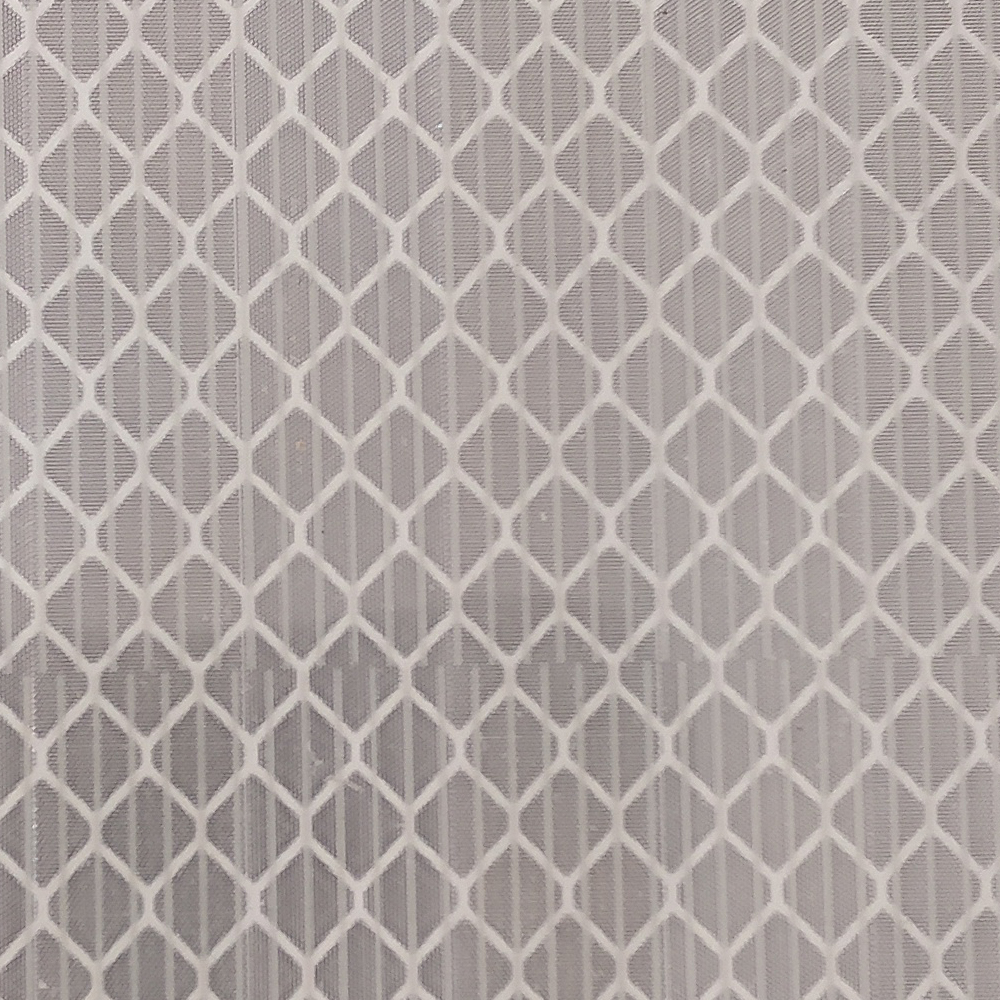} }}%
        \caption{ASTM D4956-17 \cite{ASTM_D4956-17} Types of Retroreflective Signs  }%
    \label{fig:ASTMtype}
\end{figure}

\begin{figure}[!h]%
    \centering
    \subfloat[This visualization shows a LiDAR point cloud in blue, and a detected sign rainbow colored by intensity]{{\frame{\includegraphics[width=0.2 \textwidth]{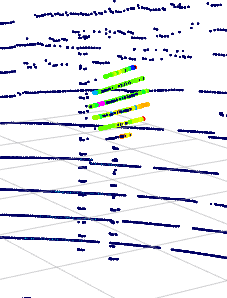} }}}%
    \qquad
    \subfloat[Typical Stop Sign on Texas A\&M's campus]{{\includegraphics[width=0.2\textwidth]{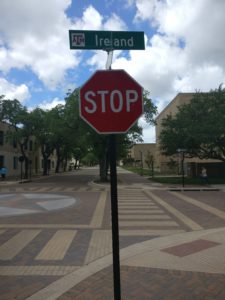} }}%
    \caption{A LiDAR visualization (a), and a camera photo (b) of a Stop Sign}%
    
    \label{fig:RVIZ_Lidar_Sign}%
\end{figure}

Street Signs in the United States are required by regulation to be made with a retroreflective sheeting ~\cite{MUTCD}, allowing for extended visibility during night-time driving. The retroreflective guidelines are set in Section 2A.08 of The Manual on Uniform Traffic Control Devices (MUTCD) \cite{MUTCD}. In these guidelines, ASTM D4956-17 \cite{ASTM_D4956-17} details several types of reflective coating. Some examples of the different types are illustrated in Figure \ref{fig:ASTMtype}, while in Figure \ref{fig:RVIZ_Lidar_Sign}, a retroreflective stop sign is visualized in both LiDAR and vision output.
The Velodyne VLP-16 and other LiDARs typically output the retro-intensity of the points, which is a measure of the ratio between the emitted energy from the LiDAR, and the returned energy from the object \cite{Georgia}. Similarly to the headlights on a car, the incident LiDAR beams on a street sign is redirected back to the LiDAR. Although measured intensity values are relative quantities, and specific to the LiDAR model/calibration, the intensities for LiDAR points on a street sign will generally be much higher than the ambient environment. Hence, LiDAR is a useful and applicable sensor for sign detection.

\subsection{Algorithm Development}
\begin{figure}[h]
\includegraphics[width=0.5\textwidth]{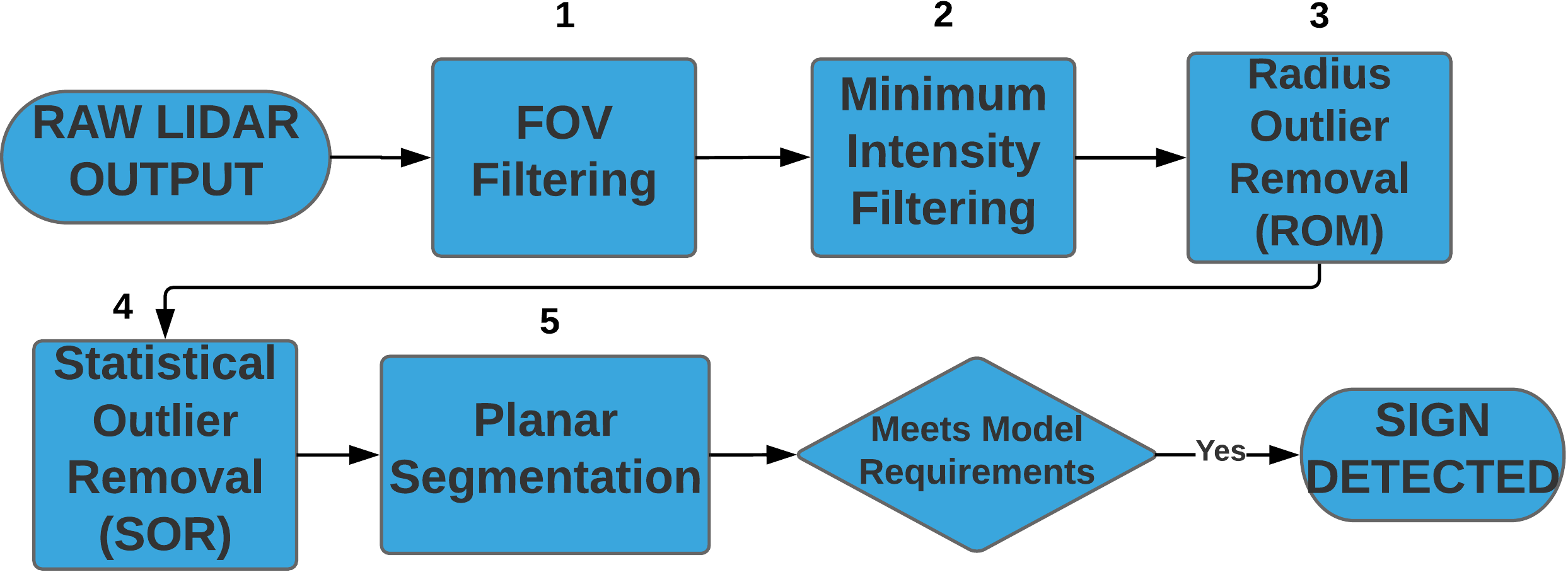}
\caption{A processing flow illustrating five techniques used to filter LiDAR output for sign detection}
\label{lidar_filter_flow}
\end{figure}

LiDAR data or LiDAR \textit{point clouds} require several processing stages to extract incident points on a sign. Five point cloud filtering stages are implemented using the Point Cloud Library (PCL) \cite{Rusu_ICRA2011_PCL} to develop a real-time algorithm that discriminates ASTM ASTM D4956-17 \cite{ASTM_D4956-17} retro-reflective street signs from a Velodyne VLP-16 LiDAR. These processes, shown in Figure \ref{lidar_filter_flow}, returns a filtered point cloud which is then later used to calculate the distance to the sign.

\subsubsection*{Stage 1 - FOV Filtering}
In this stage, an elementary Field of View (FOV) filter is applied to the dataset to reduce 50\% of the point cloud. This process step removes any data behind the LiDAR, and any data outside a $\pm$ 10 meter side clearance . Performing initial geometry based filtering alleviates computational expenses in subsequent processes. 

\subsubsection*{Stage 2 - Minimum Intensity Filtering}
The next processing step applies a minimum intensity threshold on the dataset to further reduce the point cloud. Again. measured intensity values are relative to the LiDAR model/calibration and thus, calibration is necessary for determining a suitable minimum intensity value. Signs tested with the VLP-16 included Types I-IV specified by ASTM D4956 \cite{ASTM_D4956-17} in the United States. An effective minimum intensity value of 85 was determined for day time operation.

\subsubsection*{Stage 3 - Radius Outlier Removal}
Returned points for a sign should be in close proximity within another; stop signs typically have a diameter of 0.75 m \cite{mutcd_2012}. This stage performs implements a PCL function, \textit{RadiusOutlierRemoval} \cite{Rusu_ICRA2011_PCL} which iteratively removes points that contains less than 3 neighbors in a 0.5 m radius. 

\subsubsection*{Stage 4 - Statistical Outlier Removal}
Stage four implements a PCL function, \textit{StatisticalOutlierRemoval} \cite{Rusu_ICRA2011_PCL}, to reduce further data noise and random scatter from LiDAR output. 

\subsubsection*{Stage 5 - Planar Segmentation}
The last process also implements a PCL function, \textit{SACSegmentation} \cite{Rusu_ICRA2011_PCL}, which returns the indices of inlier points that exists on plane models within allowable tolerances. The function uses a RANSAC method to segment points, and coefficients of the plane model $\textbf{n} = (a,b,c)$ is also returned \cite{Rusu_ICRA2011_PCL}. The assumption made during this process is that any points that exist along the street sign should lay on a plane.

A final check is then performed that evaluates the sign direction. Since concern is only for signs facing the vehicle, the normal facing coefficient in the plane model $ax + by + cz +d = 0$ is checked to be within some tolerance of $a \geq 0.9$. 

\subsection{Results}
\begin{figure}[]
\centering
\includegraphics[width=0.5\textwidth]{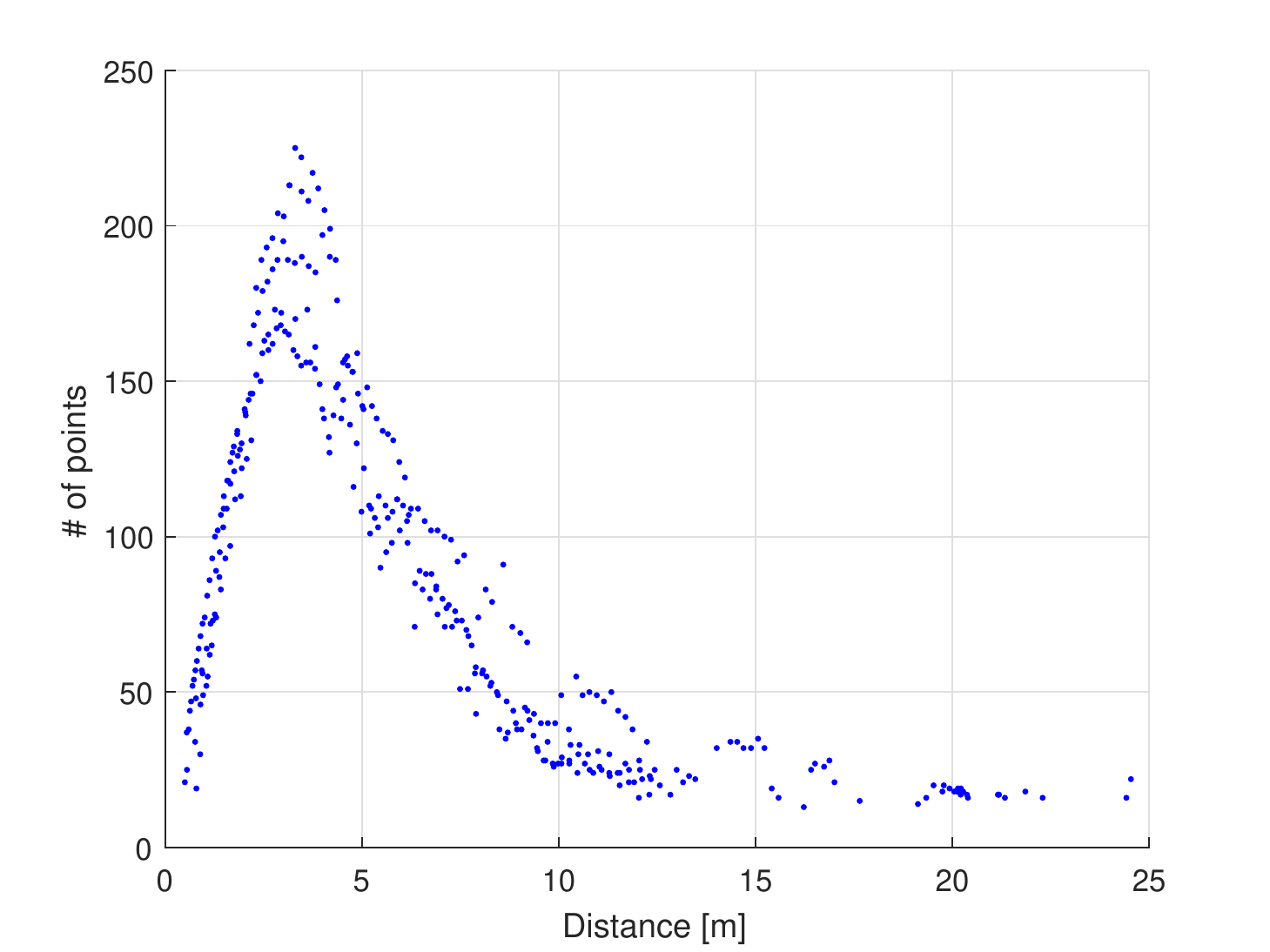}
\caption{Distribution of number of points captured vs. detection distance for sign detection}
\label{fig:signdist}
\end{figure}

\begin{figure}[!h]%
    \centering
    \subfloat[Distance = 9.2 meters, \# points = 44]{{\includegraphics[width=0.21 \textwidth]{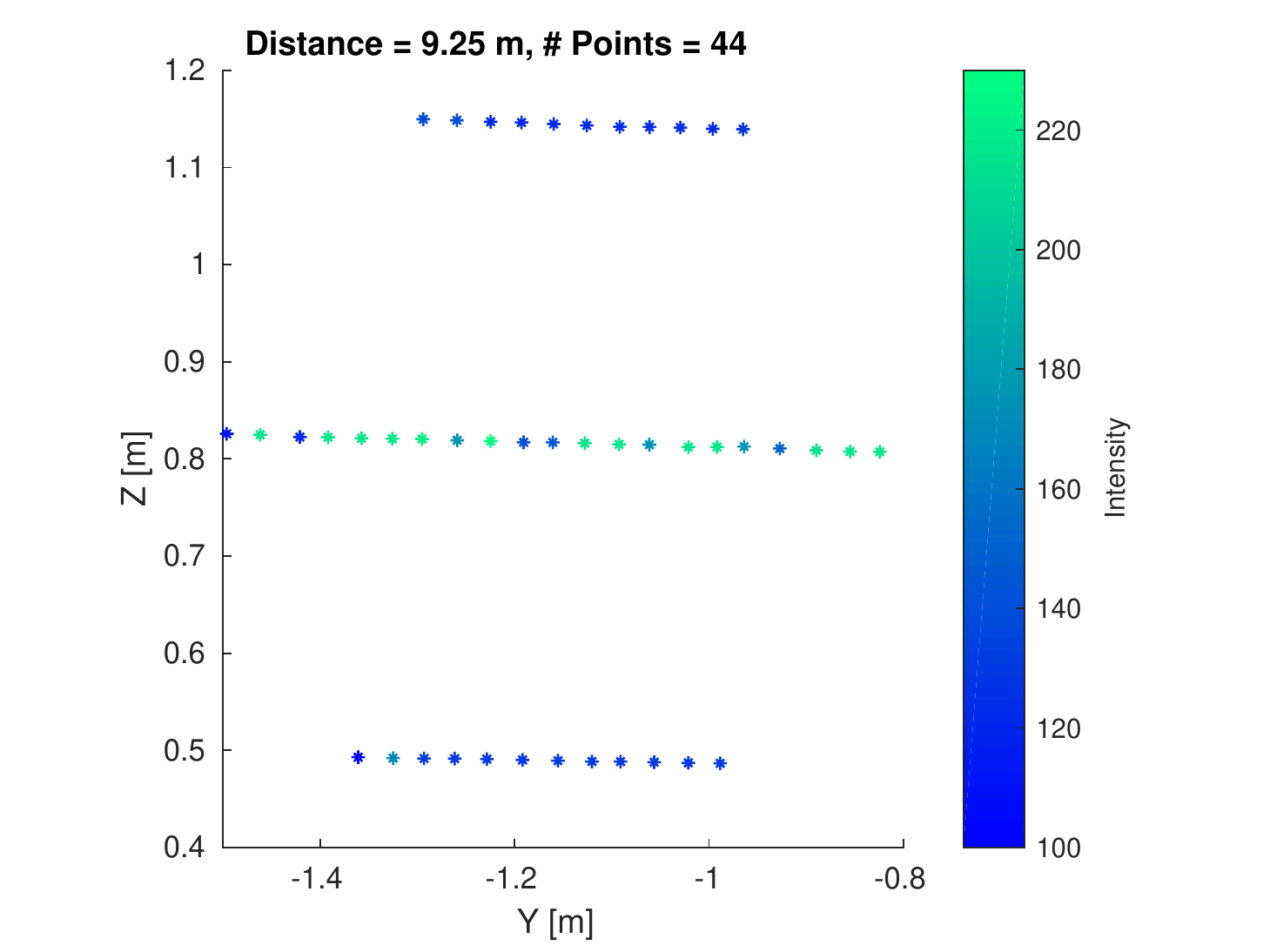} }}%
    \qquad
    \subfloat[Distance = 6.9 meters, \# points = 83]{{\includegraphics[width=0.2\textwidth]{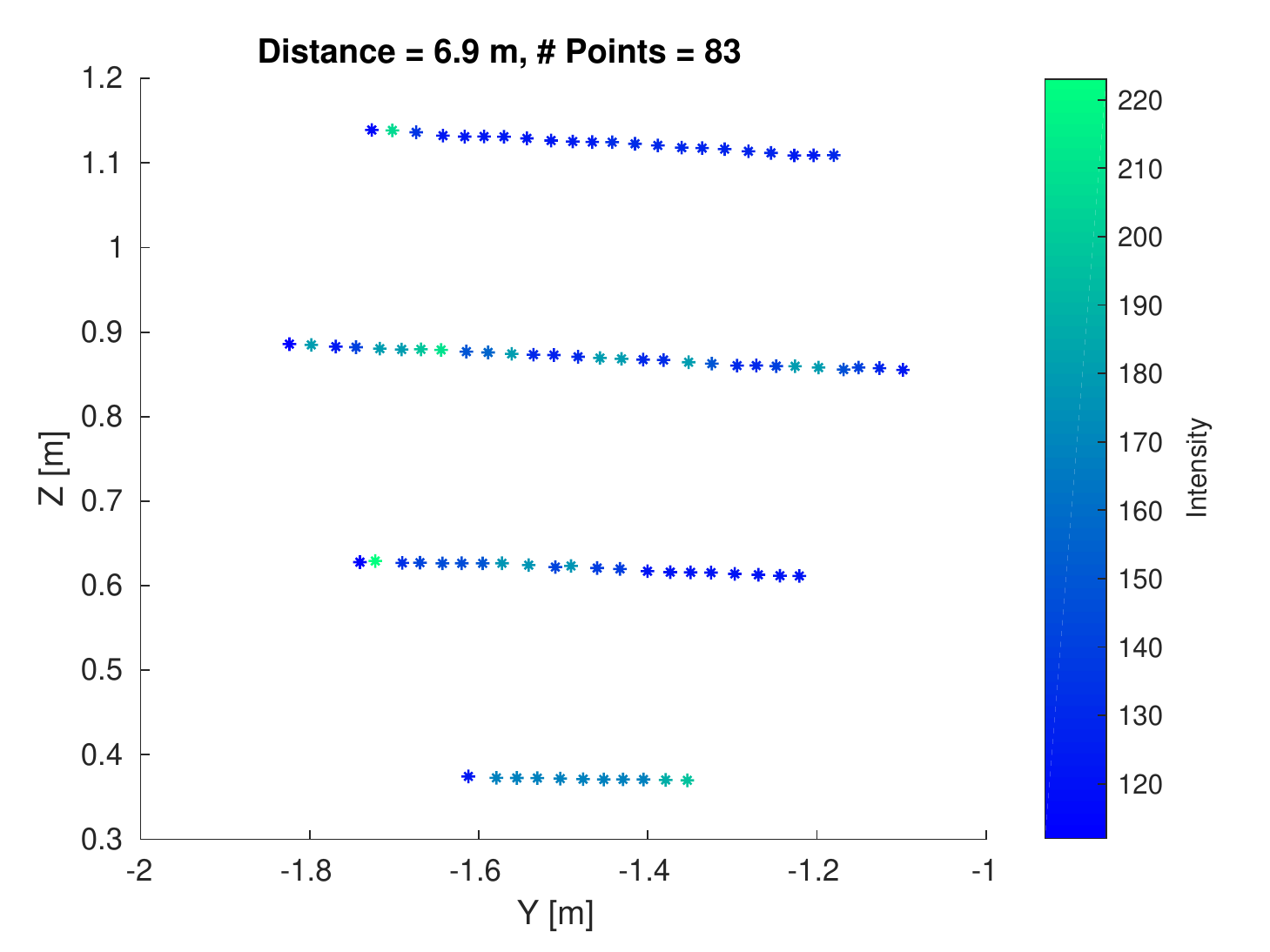} }}%
    \caption{Filtered LIDAR output at different detection distances for a stop sign}%
    \label{fig:stopsign_plot}%
\end{figure}

Figure \ref{fig:signdist} illustrates the number of LiDAR data points captured at various distances from experimentation runs during daytime, where travel speeds were limited to 3 m/s.  Signs were able to be detected from up to 20 meters away; however, only at from 10 m or \textit{N} $\approx$ 40 points were signs reliably detected. Illustrated in \ref{fig:stopsign_plot} are example plots of stop signs detected at two distances, one at 9.2 m or \textit{N} = 44 points, and another at 6.9 m or \textit{N} = 83 points. 

In both plots of Figure \ref{fig:stopsign_plot}, a hexagonal shape is captured rather than a octagon for stop signs. This down sample is a limitation imposed by the ring separation distance of the VLP-16. 

With a detected and classified stop or yield sign, the shuttle should decelerate at a rate of $a = -V_i^2/2d$ where $V_i$ is the shuttle velocity at detection, and $d$ is the distance at detection. 
With a sign detection distance of $d \approx$ 10 m, and travel speeds limited to $V_i \approx 3 m/s$,  the shuttle could comfortably decelerate at an average rate of $a = 0.15$ m/$s^2$.

Future work of this sign detection program includes: (1) event handling in cases where multiple signs exist, (2) sensor fusion with camera input, and (3) speed based handling. Currently the algorithm returns only the nearest sign, which can limit the detection distance for additional signs ahead in the trajectory. Sensor fusion with camera vision is necessary for sign classification, and also adds robustness in detection rates \cite{Georgia}. Similar to the obstacle avoidance program, the sign detection program should calculate an adjusted speed based on any detected and classified signs. This adjusted speed would then tie into the speed selector block shown in Figure \ref{system_arch}.

\section{Pedestrian Communication}
\begin{figure}[h!]
\includegraphics[width=0.5\textwidth]{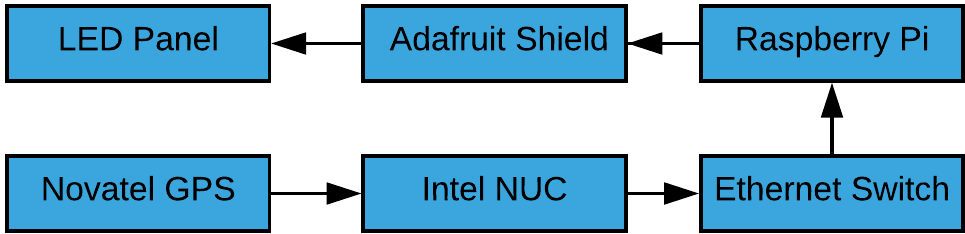}
\caption{LED panel setup}
\label{led_setup}
\end{figure}

In college campuses or other situations where a high density of pedestrians are present, it is necessary to have a way to communicate with them when there is no driver in the shuttle. When the drivers control their shuttles, they can communicate with pedestrians and other vehicles using things like hand gestures and eye contact. One solution is to provide a screen to display messages where the pedestrians can read and understand what the shuttle is doing and is going to do. The hardware setup is straightforward. As shown in Figure \ref{led_setup}, a LED screen was installed behind the windshield of the shuttle. An Adafruit RGB Matrix HAT, which was mounted on the back of the panel, was used control the LED screen \cite{Adafruit_hat}. A Raspberry Pi, which was connected to the RGB Matrix HAT, was used to generate images and send them to the LED screen.

\begin{figure}[h!]
\includegraphics[width=0.5\textwidth]{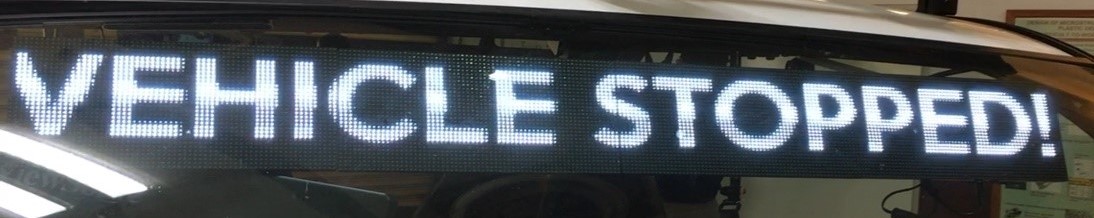}
\caption{message on the LED screen}
\label{led_messages}
\end{figure}

A ROS node running on the Raspberry Pi was used to read the current speed of the shuttle. Based on the current speed of the shuttle, the LED screen will display texts that inform whether the car is moving or not (Figure \ref{led_messages}).  

In conclusion, a display is an viable alternative solution to a driver where effectively communicating with pedestrians is crucial. The setup was able to display the appropriate messages that let the pedestrians know what the shuttle is doing. In the future, we will develop algorithms to detect pedestrians, and the shuttle will display the appropriate messages when pedestrians are present. 

\section{Conclusion \& Future Work}
Autonomous shuttles provide a good solution to the last mile problem. Their ability to drive on pedestrian pathways is considerably valuable in getting passengers as close to the destination as possible. This comes with challenges however, since these paths have dense pedestrian traffic, and are narrow with no lane markings. 

The shuttle presented in this paper is capable of operating in these conditions. Waypoint following is repeatable enough to follow narrow pedestrian paths.Effective obstacle detection and avoidance allows operation in dense traffic. Sign detection enables the shuttle to stop at intersections, and mounted signs and speakers allow it to communicate with pedestrians or other traffic.

Improvements to each section including the system architecture are planned. Local path planning will reduce the reliance on GPS in areas with bad reception. Obstacle detection can be improved by use of a better performing algorithm, and additional sensors can be used to eliminate detection blind spots. Future studies will include use of the shuttle for paratransit on a college campus, and operation and fleet management of multiple shuttles. 

\printbibliography

\end{document}